# Short Term Load Forecasting Using Multi Parameter Regression

Mrs. J. P. Rothe        Dr. A. K. Wadhwani        Dr. Mrs. S. Wadhwani

*Abstract*
**Short Term Load forecasting in this paper uses input data dependent on parameters such as load for current hour and previous two hours, temperature for current hour and previous two hours, wind for current hour and previous two hours, cloud for current hour and previous two hours. Forecasting will be of load demand for coming hour based on input parameters at that hour. In this paper we are using multiparameter regression method for forecasting which has error within tolerable range. Algorithms implementing these forecasting techniques have been programmed using MATLAB and applied to the case study. Other methodologies in this area are ANN, Fuzzy and Evolutionary Algorithms for which investigations are under process. Adaptive multiparameter regression for load forecasting, in near future will be possible.**

*Keywords*: *Weather parameters, regression, Load forecasting*.

## I. INTRODUCTION

The quality of the short-term hourly load forecasting with lead times ranging from one hour to several days ahead has a significant impact on the efficiency of operation of any electrical utility. Many operational decisions such as economic scheduling of the generating capacity, scheduling of fuel purchase and system security assessment are based on such forecasts.

Short-term weather variability and, in the longer term, climate variability has a major impact on the generation, transmission and demand for electricity. Preregulation, utilities generally managed to limit weather and climate impacts through central planning and vertical integration. In the long term, sufficient plant was planned and constructed to meet anticipated peak demand, whilst the costs arising from short-term variability were absorbed and passed on to consumers. With deregulation market participants are becoming exposed to these effects. Hence, there is greater need to manage the impact of weather and climate uncertainty.

Weather is defined as the atmospheric conditions existing over a short period in a particular location. It is often difficult to predict and can vary significantly even over a short period. Climate also varies in time: seasonally, annually and on a decadal basis [1].

Weather components such as temperature, wind speed, cloud cover play a major rule in short-term load forecasting by changing daily load curves. The characteristics of these components are reflected in the load requirements although some are affected more than others. Therefore temperature, wind speed, and cloud cover are used as variables with historical load date to improve predict hourly load.

Load demand depends on parameters such as changes in ambient temperature, wind speed, humidity, precipitation and cloud cover. As electricity demand is closely influenced by these climatic parameters, there is likely to be an impact on demand patterns. Every hour load demand depends on these crucial role parameters.

As developing countries improve their standard of living, their use of air conditioning and other weather-dependent consumption may increase the demand sensitivity to climate change.

This paper attempts at first hand statistical regression methods for existing studies on parameter impacts on electricity demand and outlines how this is being assessed for the rapidly growing Indian electricity sector.

## II. MULTIPLE REGRESSION USING MATLAB

When *y* is a function of more than one predictor variable, the matrix equations that express the relationships among the variables must be expanded to accommodate the additional data. This is called *multiple regression*.

Suppose you measure a quantity *y* for several values of $x_1$ and $x_2$. Enter these variables in the MATLAB Command Window, as follows:

x1 = [.2 .5 .6 .8 1.0 1.1]';

x2 = [.1 .3 .4 .9 1.1 1.4]';

y = [.17 .26 .28 .23 .27 .24]';

A model of this data is of the form

$$y = a_0 + a_1 x_1 + a_2 x_2$$

Multiple regression solves for unknown coefficients $a_0$, $a_1$, and $a_2$ by minimizing the sum of the squares of the deviations of the data from the model (least-squares fit).

Construct and solve the set of simultaneous equations by forming a design matrix, X, and solving for the parameters by using the backslash operator:

X = [ones (size(x1)) x1  x2];

a = X\y

a =0.1018

   0.4844

  -0.2847

The least-squares fit model of the data is

$$y = 0.1018 + 0.4844 x_1 - 0.2847 x_2$$

To validate the model, find the maximum of the absolute value of the deviation of the data from the model:

Y = X*a;

MaxErr = max (abs(Y - y))

MaxErr = 0.0038

This value is much smaller than any of the data values, indicating that this model accurately follows the data.

### III. DAILY LOAD PROFILES

The effect of changing climate may be indicated to some extent by considering the effect of changes in previous hour load demand and parameters of climate at those hours. The simplest approach to forecast is to perform a multi variable regression between coming hour load demand level and other parameters.

For one variable dependency such as demand as a function of temperature:

Di = b1 + b2 * Ti

Where Di is electrical demand at hour t, Ti is temperature at that hour t. The scatter plot and trend lines are as shown. While these clearly indicate a relationship between peak demand and temperature, the relatively low coefficient of determination R, it requires a more sophisticated approach.

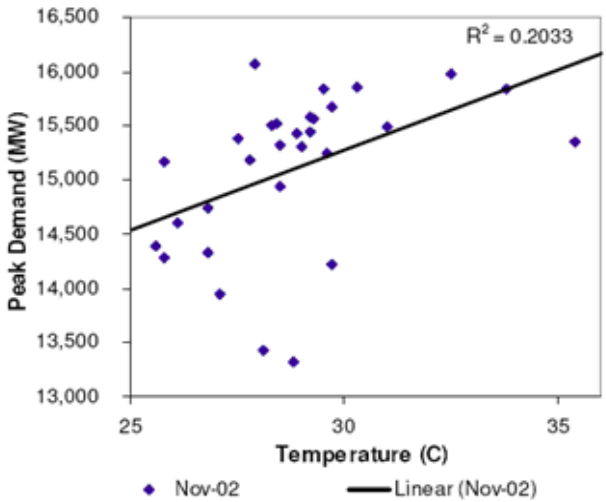

Fig. 1 Regression between temperature and demand

Multiple linear regression (MLR) is one such technique and is the most commonly used short-term forecasting method. This is due to its flexibility, as it can be used as a stand-alone method for load forecasting. At its most basic, the analysis will use forecast temperatures as well as other variables such as electrical demand in the previous hour (shown schematically in Figure 1).

The least-squares fit model of the data related to multiparameter regression for demand "y" and input variables let us say x1, x2.

$$y = 0.1018 + 0.4844 x_1 - 0.2847 x_2$$

To validate the model, find the maximum of the absolute value of the deviation of the data from the model (MATLAB COMMAND):

Y = X*a;

MaxErr = max (abs(Y - y))

MaxErr = 0.0038

This value is much smaller than any of the data values, indicating that this model accurately follows the data.

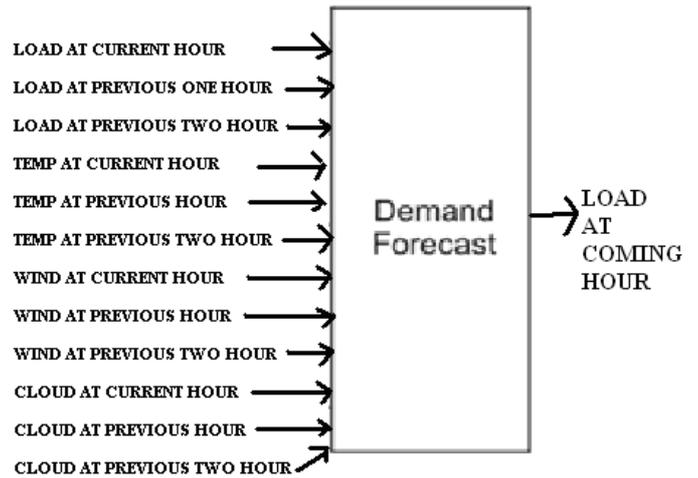

Fig. 2 Schematic STLF Model

Steps for Carrying Multiparameter Regression

- The program ("runme1.m") works using load data for 99 sets of observed values (along with temp, wind and cloud parameters at current and previous two hours). [Data is collected from Maharashtra, MSEB Kalwa].

- Multiple regression tries to fit an equation for forecasting load at coming hour based on 12 parameters using matlab procedure for regression.

- Data in text form is successfully written using a program "runme2.m".

- This multiparameter regressed equation can be used for forecasting based on any new set of four parameters (load, temp, wind and cloud) at coming hour, using program "runme3.m".

Following Graph1 and Graph2 summarizes the above work

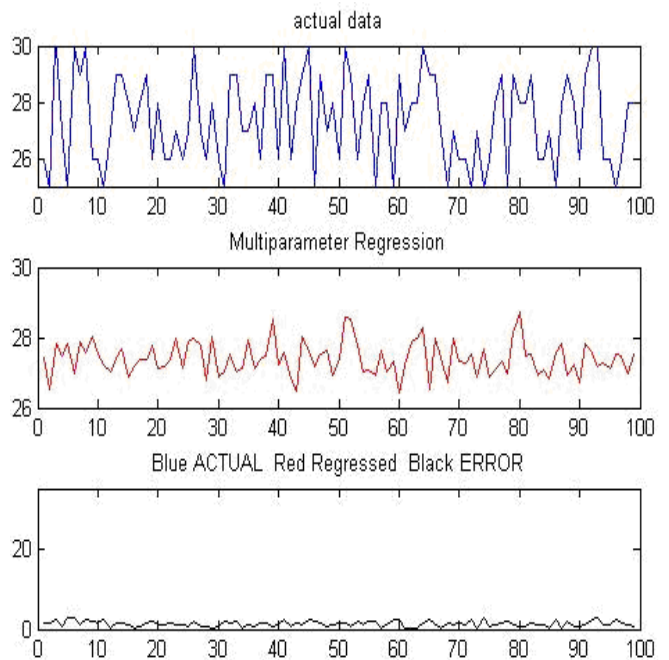

Fig. 3 Graph 1

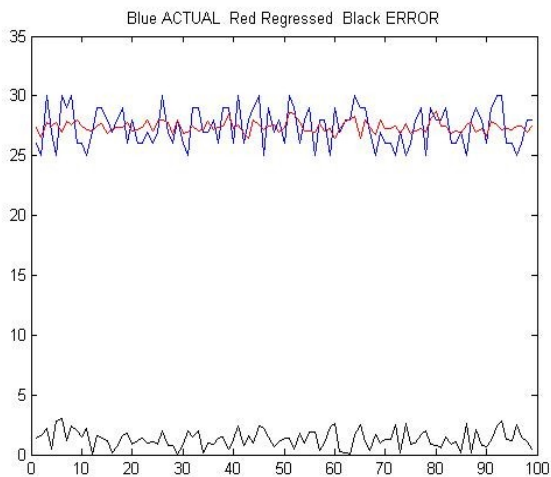

Fig. 4 Graph 2

(Input parameters for coming hour)

| Input Data |
|---|
| 29 |
| 25 |
| 28 |
| 43 |
| 41 |
| 42 |
| 30 |
| 27 |
| 28 |
| 71 |
| 71 |
| 75 |

Forecasted load (in MW) = 26.8942

TABLE I    Input Data (12 parameters)

| | | | | | | | | | |
|---|---|---|---|---|---|---|---|---|---|
| 26.00 | 26.00 | 26.00 | 30.00 | 25.00 | 25.00 | 27.00 | 27.00 | 30.00 | 27.00 |
| 30.00 | 28.00 | 25.00 | 26.00 | 29.00 | 25.00 | 28.00 | 29.00 | 25.00 | 30.00 |
| 25.00 | 26.00 | 27.00 | 26.00 | 25.00 | 30.00 | 25.00 | 25.00 | 26.00 | 27.00 |
| 45.00 | 40.00 | 45.00 | 43.00 | 42.00 | 43.00 | 44.00 | 42.00 | 43.00 | 44.00 |
| 40.00 | 40.00 | 41.00 | 44.00 | 43.00 | 42.00 | 45.00 | 44.00 | 44.00 | 44.00 |
| 42.00 | 41.00 | 41.00 | 40.00 | 44.00 | 41.00 | 42.00 | 43.00 | 44.00 | 45.00 |
| 30.00 | 26.00 | 27.00 | 27.00 | 30.00 | 30.00 | 27.00 | 30.00 | 27.00 | 28.00 |
| 28.00 | 29.00 | 29.00 | 28.00 | 27.00 | 29.00 | 26.00 | 29.00 | 25.00 | 28.00 |
| 29.00 | 28.00 | 26.00 | 26.00 | 25.00 | 28.00 | 26.00 | 28.00 | 27.00 | 28.00 |
| 72.00 | 74.00 | 73.00 | 73.00 | 71.00 | 74.00 | 71.00 | 72.00 | 74.00 | 74.00 |
| 75.00 | 72.00 | 75.00 | 70.00 | 72.00 | 74.00 | 71.00 | 72.00 | 71.00 | 73.00 |
| 70.00 | 72.00 | 72.00 | 73.00 | 70.00 | 74.00 | 73.00 | 73.00 | 75.00 | 73.00 |

TABLE II ACTUAL and STATISTICAL PREDICTED LOAD IN MW at Coming Hour

| ACTUAL PREDICTED LOAD (MW) | STATISTICAL PREDICTED LOAD ( MW) | |
|---|---|---|
| 26.00 | [1] | 27.41 |
| 25.00 | | 26.54 |
| 30.00 | | 27.81 |
| 27.00 | [2] | 27.46 |
| 25.00 | [3] | 27.81 |
| 30.00 | | 26.99 |
| 29.00 | [4] | 27.86 |
| 30.00 | [5] | 27.59 |
| 26.00 | [6] | 28.03 |
| 26.00 | [7] | 27.53 |

TABLE III    Output (forecasting)

V. CONCLUSION

Load demand at any hour may depend technically on few parameters but as far as variability of data is concerned certain other dummy variables play important role in trends rather than a technical effect. Tendencies and trends on hour to hour base based on uncontrollable parameters require to fit a more generous model of multi variable analysis technique such as MLR.

The response of such techniques to random data is appreciable since errors involved are less than 5 %. For less fluctuating trends it is bound to give accurate results by forecasting close to actual.

This paper has reviewed existing research on climate change impacts on electricity demand and has outlined how this can be assessed for forecasting in INDIA where the rapidly growing economy adds an additional level of complexity.

IV. FUTURE WORK

The research outlined above indicates the general requirements of the climate change impact programme. Further work will include:

- ANN based analysis of weather-sensitivity of electricity demand,
- Adaptive approach towards multiple regression models.